\documentclass[11pt,a4paper]{article}
\usepackage[hyperref]{emnlp2018}
\usepackage{times}
\usepackage{latexsym}

\usepackage{url}

\aclfinalcopy 

\usepackage{amsmath}
\usepackage{boldline}
\usepackage{graphicx}
\graphicspath{ {figs/} }

\usepackage{lipsum}

\newcommand\blfootnote[1]{%
  \begingroup
  \renewcommand\thefootnote{}\footnote{#1}%
  \addtocounter{footnote}{-1}%
  \endgroup
}

\def\secref#1{Sec.~\ref{#1}}

\def\eqnref#1{Eqn.~\ref{#1}}

\title{Game-Based Video-Context Dialogue}

\author{Ramakanth Pasunuru \and Mohit Bansal \\
  UNC Chapel Hill \\
  {\tt \{ram, mbansal\}@cs.unc.edu} \\
 }

\date{}
\begin{document}
\maketitle

\begin{abstract}
Current dialogue systems focus more on textual and speech context knowledge and are usually based on two speakers. Some recent work has investigated static image-based dialogue. However, several real-world human interactions also involve dynamic visual context (similar to videos) as well as dialogue exchanges among multiple speakers. To move closer towards such multimodal conversational skills and visually-situated applications, we introduce a new video-context, many-speaker dialogue dataset based on live-broadcast soccer game videos and chats from Twitch.tv. This challenging testbed allows us to develop visually-grounded dialogue models that should generate relevant temporal and spatial event language from the live video, while also being relevant to the chat history. For strong baselines, we also present several discriminative and generative models, e.g., based on tridirectional attention flow (TriDAF). We evaluate these models via retrieval ranking-recall, automatic phrase-matching metrics, as well as human evaluation studies. We also present dataset analyses, model ablations, and visualizations to understand the contribution of different modalities and model components. 
\end{abstract}

\section{Introduction}

\begin{figure}[ht]
\centering
\includegraphics[width=0.9\linewidth]{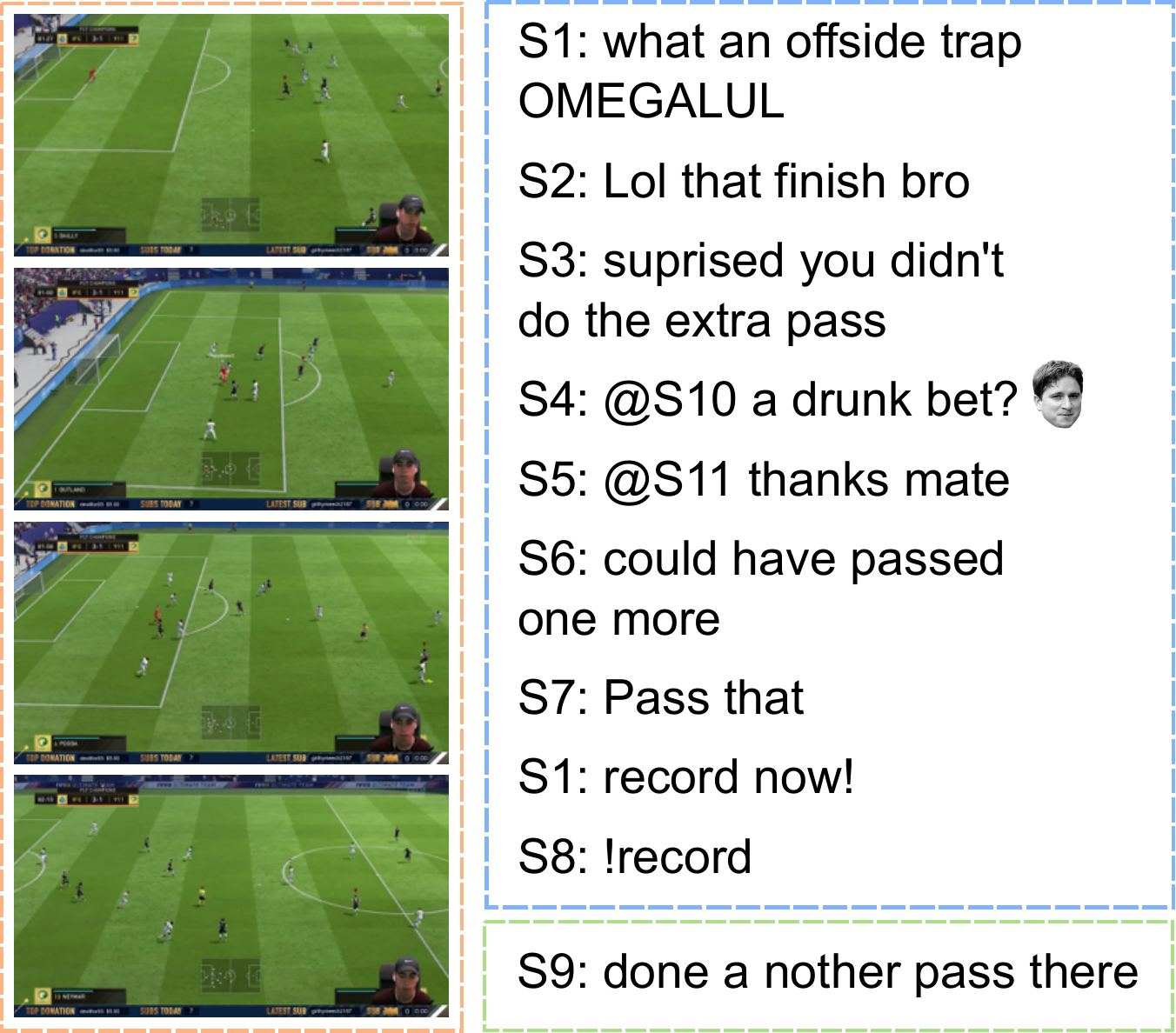}
\vspace{-6pt}
\caption{Sample example from our many-speaker, video-context dialogue dataset, based on live soccer game chat. The task is to predict the response (bottom-right) using the video context (left) and the chat context (top-right).}
\label{fig:title_figure}
\end{figure}

Dialogue systems or conversational agents which are able to hold natural, relevant, and coherent interactions with humans have been a long-standing goal of artificial intelligence and machine learning.\blfootnote{We  release all data, code, and models at: {\scriptsize\url{https://github.com/ramakanth-pasunuru/video-dialogue}}} There has been a lot of important previous work in this field for decades~\cite{weizenbaum1966eliza,isbell2000cobot,rambow2001natural,rieser2005corpus,georgila2006user,rieser2008learning,ritter2011data}, including recent work on introduction of large textual-dialogue datasets (e.g.,~\newcite{lowe2015ubuntu,serban2016building}) and end-to-end neural network based models~\cite{sordoni2015neural,vinyals2015neural,su2016line,luan2016lstm,li2016persona,serban2017multiresolution,serban2017hierarchical}.

Current dialogue tasks are usually focused on the textual or verbal context (conversation history). In terms of multimodal dialogue, speech-based spoken dialogue systems have been widely explored~\cite{eckert1997user,singh2000reinforcement,young2000probabilistic,janin2003icsi,deep-learning-spoken-text-dialog-systems,wen2015semantically,su2016line,mrkvsic2016neural}, as well as work on gesture and haptics based dialogue~\cite{johnston2002match,cassell1999embodied,foster2008roles}. In order to address the additional advantage of using visually-grounded context knowledge in dialogue, recent work introduced the visual dialogue task~\cite{das2016visual,de2016guesswhat,mostafazadeh2017image}. However, the visual context in these tasks is limited to one static image. Moreover, the interactions are between two speakers with fixed roles (one asks questions and the other answers).

Several situations of real-world dialogue among humans involve more `dynamic' visual context, i.e., video-style information of the world moving around us (both spatially and temporally). Further, several human conversations involve more than two speakers, with changing roles. In order to develop such dynamically-visual multimodal dialogue models, we introduce a new `many-speaker, video-context chat' testbed, along with a new dataset and models for the same.
Our dataset is based on live-broadcast soccer (FIFA-18) game videos from the `Twitch.tv' live video streaming platform, along with the spontaneous, many-speaker live chats about the game. This challenging testbed allows us to develop dialogue models where the generated response is required to be relevant to the temporal and spatial events in the live video, as well as be relevant to the chat history (with potential impact towards video-grounded applications such as personal assistants, intelligent tutors, and human-robot collaboration).

We also present several strong discriminative and generative baselines that learn to retrieve and generate bimodal-relevant responses. We first present a triple-encoder discriminative model to encode the video, chat history, and response, and then classify the relevance label of the response. We then improve over this model via tridirectional attention flow (TriDAF). For the generative models, we model bidirectional attention flow between the video and textual chat context encoders, which then decodes the response.
We evaluate these models via retrieval ranking-recall, phrase-matching metrics, as well as human evaluation studies.
We also present dataset analysis as well as model ablations and attention visualizations to understand the contribution of the video vs. chat modalities and the model components.


\section{Related Work}

Early dialogue systems had components of natural language (NL) understanding unit, dialogue manager, and NL generation unit~\cite{bates1995models}. 
Statistical learning methods were used for automatic feature extraction~\cite{dowding1993gemini,mikolov2013distributed}, dialogue managers incorporated reward-driven reinforcement learning~\cite{young2013pomdp,shah2016interactive}, and the generation units have been extended with seq2seq neural network models~\cite{vinyals2015neural,serban2016building,luan2016lstm}.

In addition to the focus on textual dialogue context, using multimodal context brings more potential for having real-world grounded conversations. For example, spoken dialogue systems have been widely explored~\cite{singh2000reinforcement,gurevych2004semantic,georgila2006user,eckert1997user,young2000probabilistic,janin2003icsi,de2007spoken,wen2015semantically,su2016line,mrkvsic2016neural,hori2016dialog,celikyilmaz2015universal,deep-learning-spoken-text-dialog-systems}, as well as gesture and haptics based dialogue~\cite{johnston2002match,cassell1999embodied,foster2008roles}. Additionally, dialogue systems for digital personal assistants are also well explored~\cite{myers2007intelligent,sarikaya2016overview,damacharla2018effects}.
In the visual modality direction, some important recent attempts have been made to use static image based context in dialogue systems~\cite{das2016visual,de2016guesswhat,mostafazadeh2017image}, who proposed the `visual dialog' task, where the human can ask questions on a static image, and an agent interacts by answering these questions based on the previous chat context and the image's visual features. Also,~\newcite{celikyilmaz2014resolving} used visual display information for on-screen item resolution in utterances for improving personal digital assistants.

In contrast, we propose to employ dynamic video-based information as visual context knowledge in dialogue models, so as to move towards video-grounded intelligent assistant applications. In the video+language direction, previous work has looked at video captioning~\cite{venugopalan2015sequence} as well as Q\&A and fill-in-the-blank tasks on videos~\cite{tapaswi2016movieqa,jang2017tgif,maharaj2016dataset} and interactive 3D environments~\cite{das2017embodied,yan2018chalet,gordon2017iqa,anderson2017vision}. There has also been early related work on generating sportscast commentaries from simulation (RoboCup) soccer videos represented as non-visual state information~\cite{chen2008learning}. Also, ~\newcite{liu2016task} presented some initial ideas on robots learning grounded task representations by watching and interacting with humans performing the task (i.e., by converting human demonstration videos to Causal And-Or graphs).
On the other hand, we propose a new video-chat dataset where the dialogue models need to generate the next response in the sequence of chats, conditioned both on the raw video features as well as the previous textual chat history.
Moreover, our new dataset presents a many-speaker conversation setting, similar to previous work on meeting understanding and Computer Supported Cooperative Work (CSCW)~\cite{janin2003icsi,waibel2001advances,schmidt1992taking}.
In the live video stream direction, ~\newcite{fu2017video} and~\newcite{ping2017video} used real-time comments to predict the frame highlights in a video, and~\newcite{barbieri2017towards} presented emotes and troll prediction.

\section{Twitch-FIFA Dataset}
\begin{figure}
\centering
\includegraphics[width=0.98\linewidth]{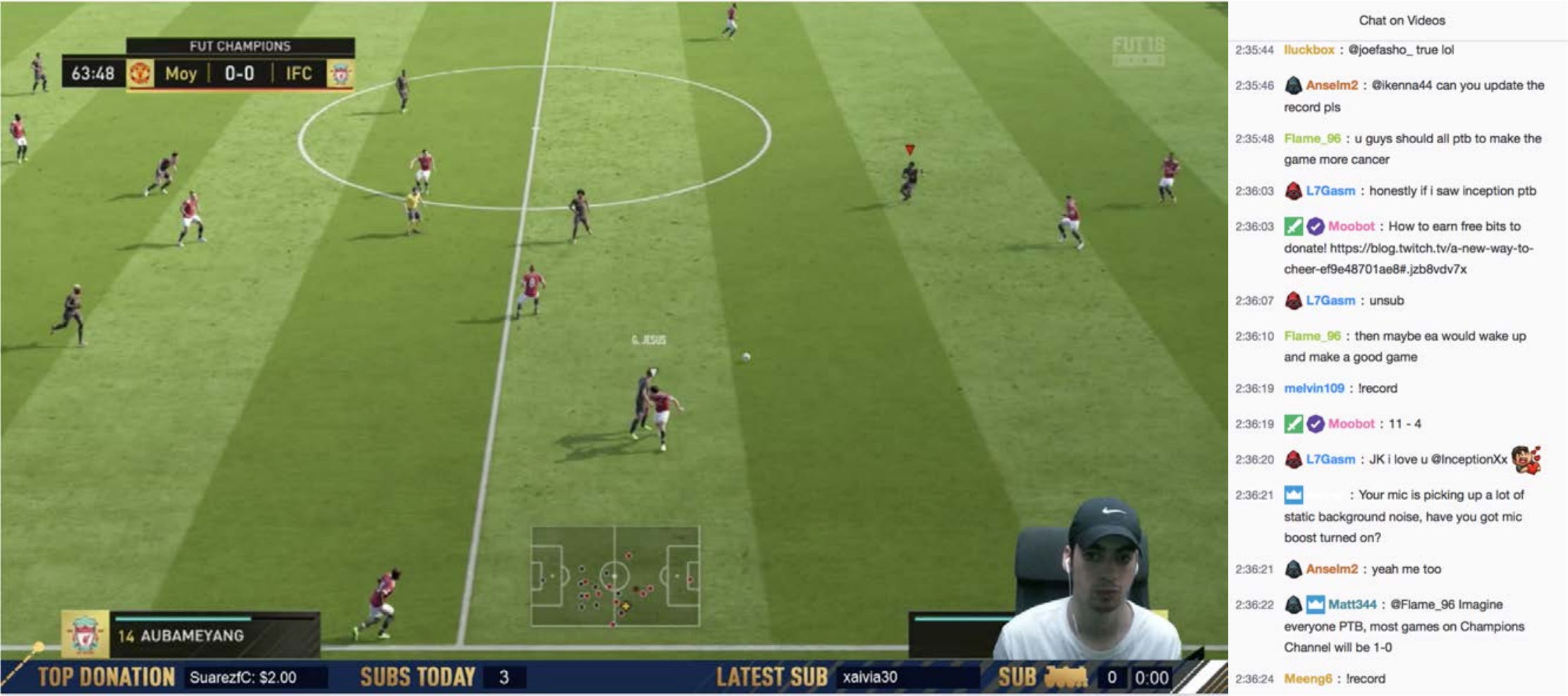}
\vspace{-10pt}
\caption{Sample page of live broadcast of FIFA-18 game on twitch.tv with concurrent user chat.\label{fig:twitch_fifa_webpage}\vspace{-10pt}}
\end{figure}

\subsection{Dataset Collection and Processing}
\label{sec:datafilter}
For our new video-context dialogue task, we used the publicly accessible Twitch.tv live broadcast platform, and collected videos of soccer (FIFA-18) games along with the users' live chat conversations about the game. This dataset has videos involving various realistic human actions and events in a complex sports environment and hence serves as a good testbed and first step towards multimodal video-based dialogue data. An example is shown in Fig.~\ref{fig:title_figure} (and an original screenshot example in Fig.~\ref{fig:twitch_fifa_webpage}), where the users perform a complex `many-speaker', `multimodal' dialogue. Overall, we collected $49$ FIFA-18 game videos along with their users' chat, and divided them into $33$ videos for training, $8$ videos for validation, and $8$ videos for testing. Each such video is several hours long, providing a good amount of data (Table~\ref{table:dataset-statistics}). 

To extract triples (instances) of video context, chat context, and response from this data, we divide these videos based on the fixed time frames instead of fixed number of utterances in order to maintain conversation topic clusters (because of the sparse nature of chat utterances count over the time). First, we use $20$-sec context windows to extract the video clips and users utterances in this time frame, and use it as our video and chat contexts, resp. Next, the chat utterances in the immediately-following $10$-sec window (response window) that do not overlap with the next instance's context window are considered as potential responses.\footnote{We use non-overlapping windows because: (1) the utterances are non-uniformly distributed in time and hence if we have a shifting window, sometimes a particular data instance/chunk becomes very sparse and contains almost zero utterances; (2) we do not want overlap between response of one window with the context of the next window, so as to avoid the encoder already having seen the response (as part of context) that the decoder needs to generate for the other window.} Hence, there are only two instances (triples) in a $60$-sec long video, i.e., $20$-sec video+chat context window and $10$-sec response window, and there is no overlap between the instances. Now, out of these potential responses, to only allow the response that has at least some good coherence and relevance with the chat context's topic, we choose the first (earliest) response that has high similarity with some other utterance in this response window (using 0.5 BLEU-4 threshold, based on manual inspection).\footnote{Based on intuition that if multiple speakers are saying the same response in that 10-second window, then this response should be more meaningful/relevant w.r.t. chat context.} 

\begin{table}[t]
\centering
\small
\begin{tabular}{l|c}
\hline
 & Relevance to Video+Chat\\
\hline
filtered response wins & 34\% \\
1st response wins & 3\%\\
Non-distinguishable & 63\% (56 both-good, 7 both-bad) \\
\hline
\end{tabular}
\vspace{-7pt}
\caption{Human evaluation of our dataset, comparing our filtered responses versus the first response in the window (for relevance w.r.t. video and chat contexts).\label{table:dataset-human-eval}\vspace{-10pt}}
\end{table}

\noindent\textbf{Human Quality Evaluation of Data Filtering Process}: To evaluate the quality of the responses that result from our filtering process described above, we performed an anonymous (randomly shuffled w/o identity) human comparison between the response selected by our filtering process vs. the first response from the response window without any filtering, based on relevance w.r.t. video and chat context. Table~\ref{table:dataset-human-eval} presents the results on $100$ sample size, showing that humans in a blind-test found $90\%$ (34+56) of our filtered responses as valid responses, verifying that our response selection procedure is reasonable. Furthermore, out of these $90\%$ valid responses, we found that $55\%$ are chat-only relevant, $11\%$ are video-only relevant, and $24\%$ are both video+chat relevant.

In order to make the above procedure safe and to make the dataset more challenging, we also discourage frequent responses (top-20 most-frequent generic utterances) unless no other response satisfies the similarity condition, hence suppressing the frequent responses.\footnote{Note that this filtering suppresses the performance of simple frequent-response baseline described in Sec.~\ref{subsec:very-simple-baselines}.} If we couldn't find any utterance based on the multi-response matching procedure described above, then we just consider the first utterance in the 10-second window as the response.\footnote{Other preprocessing steps include: omit the utterances in the response window which refer to a speaker name out of the current chat context; remove non-representative utterances, e.g., those with hyperlinks; replace (anonymize) all the user identities mentioned in the utterances with a common tag (i.e., anonymizing due to similar intuitions from the Q\&A community~\cite{hermann2015teaching}).} 
We also make sure that the chat context window has at least $4$ utterances, otherwise we exclude that context window and also the corresponding response window from the dataset. After all this processing, our final resulting dataset contains $10,510$ samples in training, $2,153$ samples in validation, and $2,780$ samples in test.\footnote{Note that this is substantially larger than or comparable to most current video captioning datasets. We plan to further extend our dataset based on diverse games and video types.}

\begin{table}
\vspace{-5pt}
\small
\centering
\begin{tabular}{l|c|c|c}
\hline
Statistics & Train & Val & Test\\
\hlineB{2}
\#Videos & 33 & 8 & 8 \\
Total Hours & 58.4 & 11.9 & 15.4 \\
Final Filtered \#Instances & 10,510 & 2,153 & 2,780 \\
Avg. Chat Context Length & 69.0 & 63.5 & 71.2 \\
Avg. Response Length & 6.5 & 6.5 & 6.1 \\
\hline
\end{tabular}
\vspace{-5pt}
\caption{Twitch-FIFA dataset's chat statistics (lengths are defined in terms of number of words).\label{table:dataset-statistics}\vspace{-7pt}}
\end{table}

\subsection{Dataset Analysis}
\label{subsec:dataset-analysis}

\noindent\textbf{Dataset Statistics}
Table~\ref{table:dataset-statistics} presents the full statistics on train, validation, and test sets of our Twitch-FIFA dataset, after the filtering process described in Sec.~\ref{sec:datafilter}. As shown, the average chat context length in the dataset is around $68$ words, and the average response length is $6.3$ words.

\begin{figure}[t]
\centering
\includegraphics[width=0.7\linewidth]{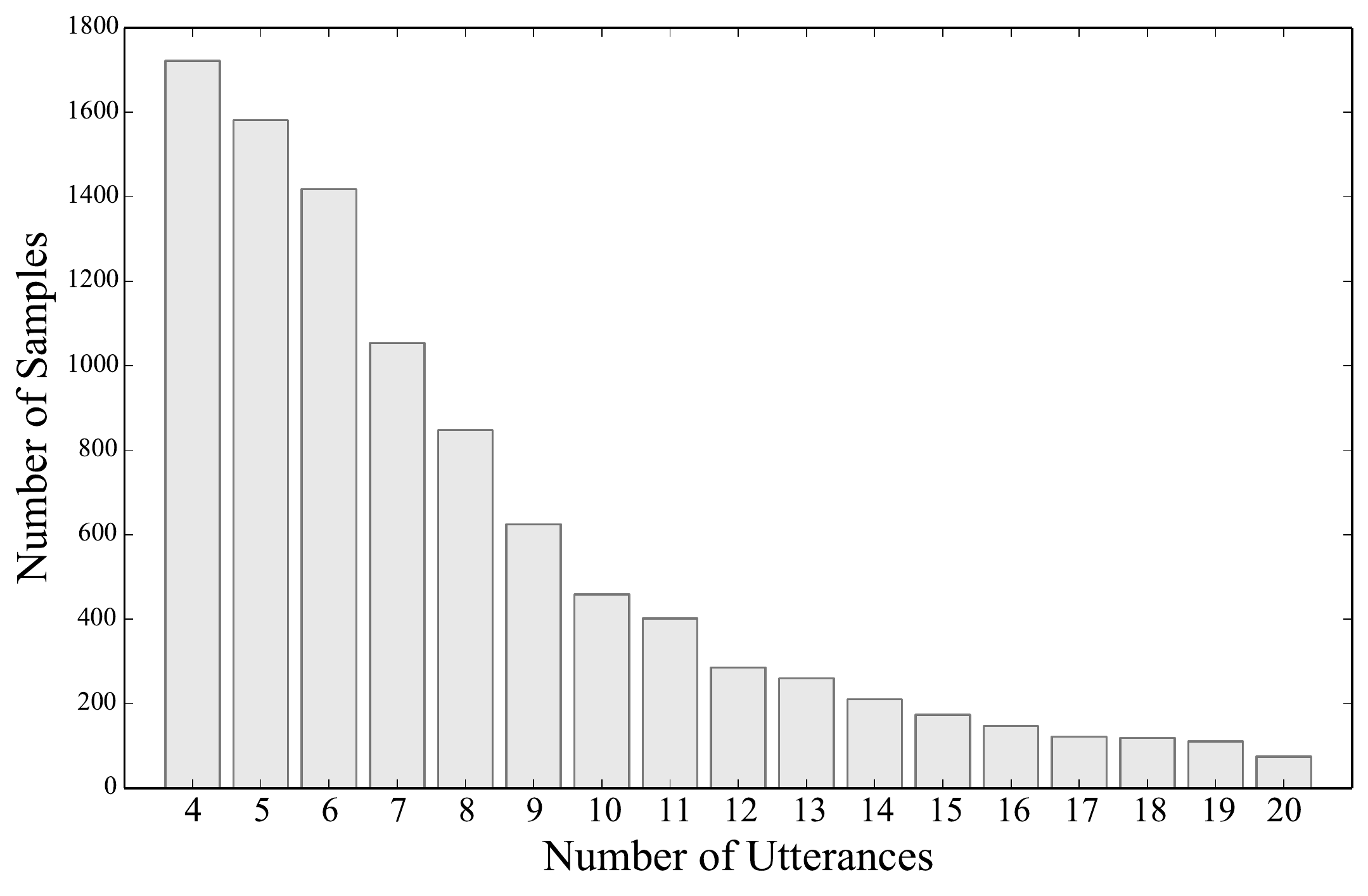}
\vspace{-7pt}
\caption{Distribution of \#utterances in chat context (w.r.t. the \#training examples for each case).
}
\label{fig:samples_vs_utterances}
\end{figure}

\noindent\textbf{Chat Context Size}
Fig.~\ref{fig:samples_vs_utterances} presents the study of number of utterances in the chat context vs. the number of such training samples. As we limit the minimum number of utterances to $4$, chat context with less than $4$ utterances is not present in the dataset. From the Fig.~\ref{fig:samples_vs_utterances}, it is clear that as the number of utterances in the chat context increases, the number of such training samples decrease. 

\begin{figure}[t]
\centering
\includegraphics[width=0.7\linewidth]{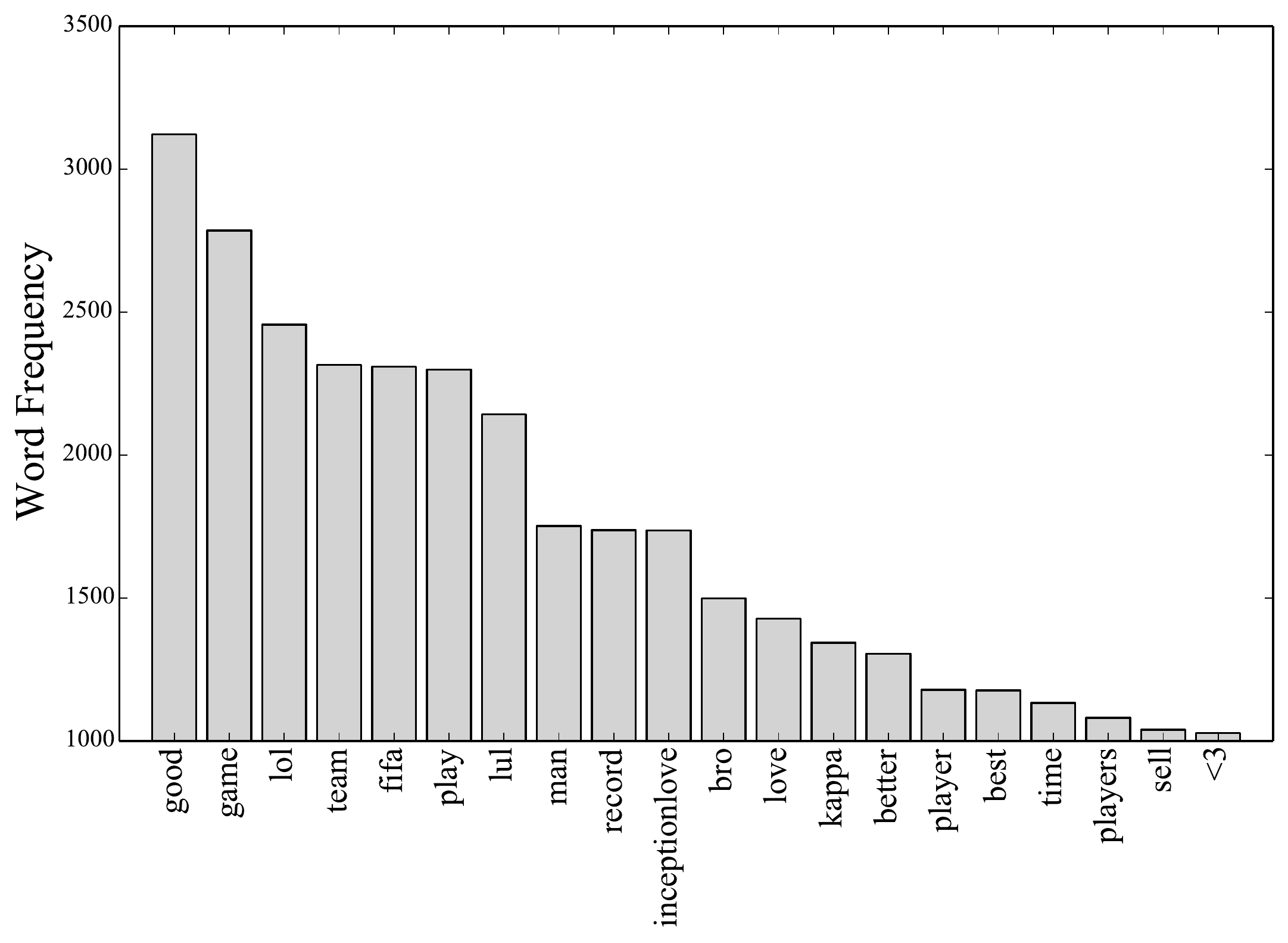}
\vspace{-10pt}
\caption{Frequent words in our Twitch-FIFA dataset.
\label{fig:frequent-words}
\vspace{-7pt}
}
\end{figure}

\noindent\textbf{Frequent Words}
Fig.~\ref{fig:frequent-words} presents the top-$20$ frequent words (excluding stop words) and their corresponding frequency in our Twitch-FIFA dataset. Most of these frequent words are related to soccer vocabulary. Also, some of these frequent words are twitch emotes (e.g. `kappa', `inceptionlove').

\section{Models}
\label{sec-methods}
Let $v=\{v_1,v_2,..,v_m\}$ be the video context frames, $u=\{u_1,u_2,..,u_n\}$ be the textual chat (utterance) context tokens, and $r=\{r_1,r_2,..,r_k\}$ be response tokens generated (or retrieved).

\subsection{Baselines}
\label{subsec:very-simple-baselines}

Our simple non-trained baselines are Most-Frequent-Response (re-rank the candidate responses based on their frequency in the training set), Chat-Response-Cosine (re-rank the candidate responses based on their similarity score w.r.t. the chat context), and Nearest-Neighbor (find the $K$-best similar chat contexts in the training set, take their corresponding responses, and then re-rank the candidate responses based on mean similarity score w.r.t. this $K$-best response set). For trained baselines, we use logistic regression and Naive Bayes methods. We use the final state of a Twitch-trained RNN Language Model to represent the chat context and response. Please see supplementary for full details.

\subsection{Discriminative Models}

\begin{figure}
\centering
\includegraphics[width=0.98\linewidth]{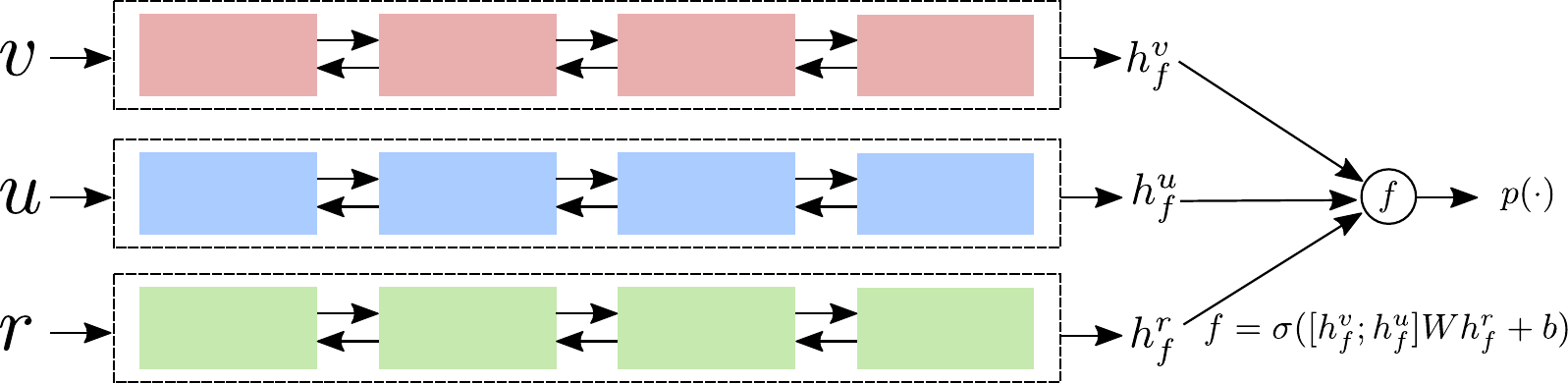}
\vspace{-5pt}
\caption{Overview of our `triple encoder' discriminative model, with bidirectional-LSTM-RNN encoders for video, chat context, and response.}
\label{fig:triple-birnn-encoder}
\vspace{-5pt}
\end{figure}

\subsubsection{Triple Encoder}
\label{subsec:triple-encoder}
For our simpler discriminative model, we use a `triple encoder' to encode the video context, chat context, and response (see Fig.~\ref{fig:triple-birnn-encoder}), as an extension of the dual encoder model in~\newcite{lowe2015ubuntu}. The task here is to predict the given training triple $(v,u,r)$ as positive or negative. Let $h^v_f$, $h^u_f$, and $h^r_f$ be the final state information of the video, chat, and response LSTM-RNN (bidirectional) encoders respectively; then the probability of a positive training triple is defined as follows:
\begin{equation}
\label{eq:sigmoid-prob}
p(v,u,r;\theta) = \sigma([h^v_f;h^u_f]^T W h^r_f + b)
\end{equation}
where $W$ and $b$ are trainable parameters. Here, $W$ can be viewed as a similarity matrix which will bring the context $[h^v_f;h^u_f]$ into the same space as the response $h^r_f$, and get a suitable similarity score. 

For optimizing our discriminative model, we use max-margin loss function similar to~\newcite{mao2016generation} and \newcite{yu2016joint}.
Given a positive training triple $(v,u,r)$, let the corresponding negative training triples be $(v',u,r)$, $(v,u',r)$, and $(v,u,r')$, i.e., one modality is wrong at a time in each of these three (see Sec.~\ref{sec:setup} for the negative example selection). The max-margin loss is:
\begin{equation}
\label{eq:max-margin-loss}
\small
\begin{aligned}
L(\theta) &= \sum [\max(0,M+ \log p(v',u,r) - \log p(v,u,r)) \\
& + \max(0,M+ \log p(v,u',r)- \log p(v,u,r)) \\
& + \max(0,M+ \log p(v,u,r')- \log p(v,u,r))]
\end{aligned}
\end{equation}
where the summation is over all the training triples in the dataset. $M$ is a tunable margin hyperparameter between positive and negative training triples.

\subsubsection{Tridirectional Attention Flow (TriDAF)}
\label{subsubsec:tri-daf}
\begin{figure}
\centering
\includegraphics[width=0.98\linewidth]{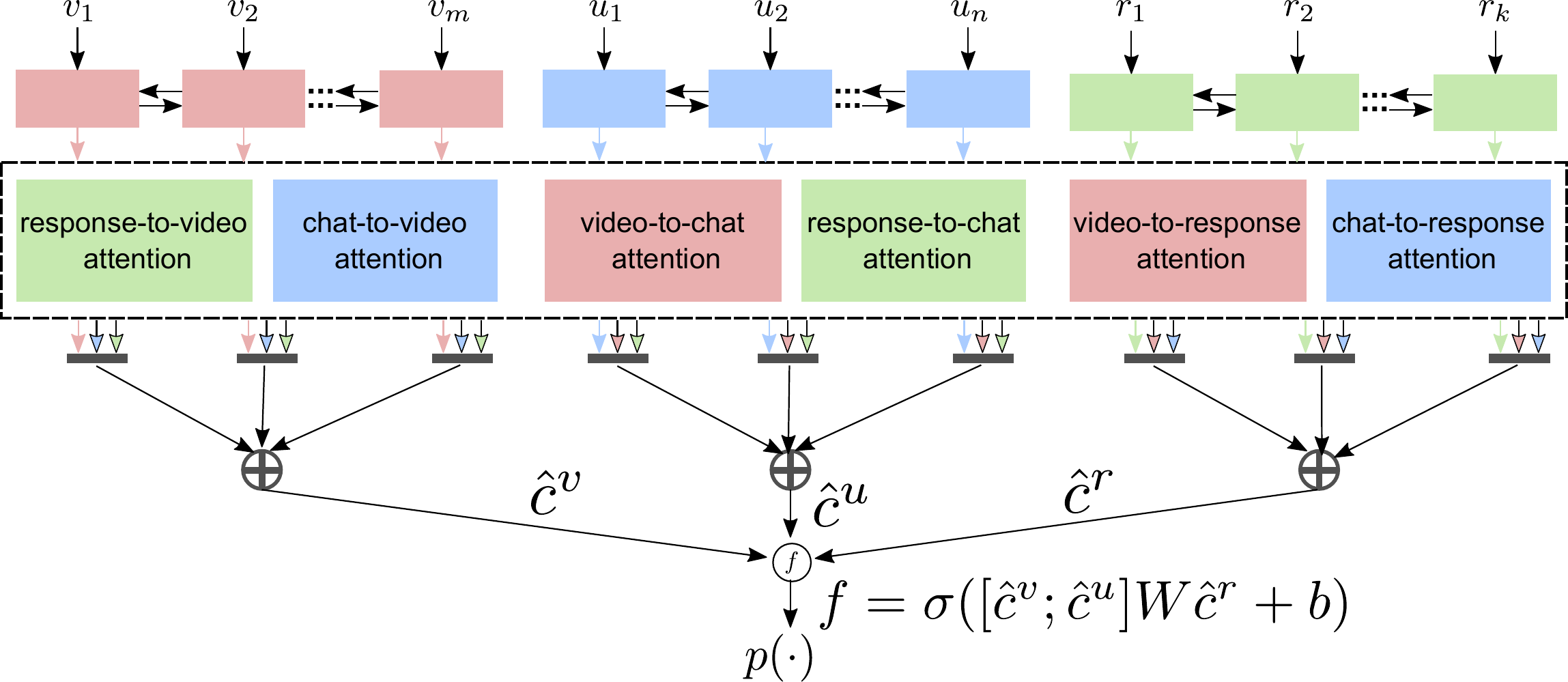}
\vspace{-7pt}
\caption{Overview of our tridirectional attention flow (TriDAF) model with all pairwise modality attention modules, as well as self-attention on video context, chat context, and response as inputs. \label{fig:tridaf-retrieval} \vspace{-7pt}
}
\end{figure}

Our tridirectional attention flow model learns stronger joint spaces between the three modalities in a mutual-information way. We use bidirectional attention flow mechanisms~\cite{seo2016bidirectional} between the video and chat contexts, between the video context and the response, as well as between the chat context and the response, hence enabling attention flow across all three modalities, as shown in Fig.~\ref{fig:tridaf-retrieval}. We name this model Tridirectional Attention Flow or TriDAF. 
We will next discuss the bidirectional attention flow mechanism between video and chat contexts, but the same formulation holds true for bidirectional attention between video context and response, and between chat context and response. Given the video context hidden state $h^v_i$ and chat context hidden state $h^u_j$ at time steps $i$ and $j$ respectively, the bidirectional attention mechanism is based on the similarity score:
\begin{equation}
S^{(v,u)}_{i,j} = w_{S^{(v,u)}}^T [h^v_i;h^u_j;h^v_i \odot h^u_j]
\end{equation}
where $S^{(v,u)}_{i,j}$ is a scalar, $w_{S^{(v,u)}}$ is a trainable parameter, and $\odot$ denote element-wise multiplication. The attention distribution from chat context to video context is defined as $\alpha_{i:}= softmax(S_{i:})$, hence the chat-to-video context vector $c^{v\leftarrow u}_i = \sum_j \alpha_{i,j}h^u_j $. 
Similarly, the attention distribution from video context to chat context is defined as $\beta_{j:} = softmax(S_{:j})$, hence the video-to-chat context vector $c^{u\leftarrow v}_j = \sum_i \beta_{j,i}h^v_i$.

We then compute similar bidirectional attention flow mechanisms between the video context and response, and between the chat context and response. Then, we concatenate each hidden state and its corresponding context vector from other two modalities, e.g., $ \hat{h}^v_i = [h^v_i; c^{v\leftarrow u}_i; c^{v\leftarrow r}_i]$ for the $i^{\text{th}}$ timestep of the video context. Finally, we add self-attention mechanism~\cite{lin2017structured} across the concatenated hidden states of each of the three modules.\footnote{In our preliminary experiments, we found that adding self-attention is $0.92\%$ better in recall@1 and faster than passing the hidden states through another layer of RNN, as done in~\newcite{seo2016bidirectional}.} If $\hat{h}^v_i$ is the final concatenated vector of the video context at time step $i$, then the self-attention weights $\alpha^s$ for this video context are the softmax of $e^s$:
\begin{equation}
e^s_i = V^v_a \tanh(W^v_a \hat{h}^v_i +b^v_a)
\end{equation}
where $V_a^v$, $W^v_a$, and $b^v_a$ are trainable self-attention parameters. The final representation vector of the full video context after self-attention is $\hat{c}^v=\sum_i \alpha^s_i \hat{h}^v_i$. Similarly, the final representation vectors of the chat context and the response are $\hat{c}^u$ and $\hat{c}^r$, respectively. Finally, the probability that the given training triple $(v,u,r)$ is positive is:
\begin{equation}
p(v,u,r;\theta) = \sigma([\hat{c}^v;\hat{c}^u]^T W \hat{c}^r + b)
\end{equation}
Again, here also we use max-margin loss (Eqn.~\ref{eq:max-margin-loss}).

\subsection{Generative Models}
\label{subsec:generative-model}

\subsubsection{Seq2seq with Attention}
\label{subsubsec:seq2seq}
\begin{figure}
\centering
\includegraphics[width=0.98\linewidth]{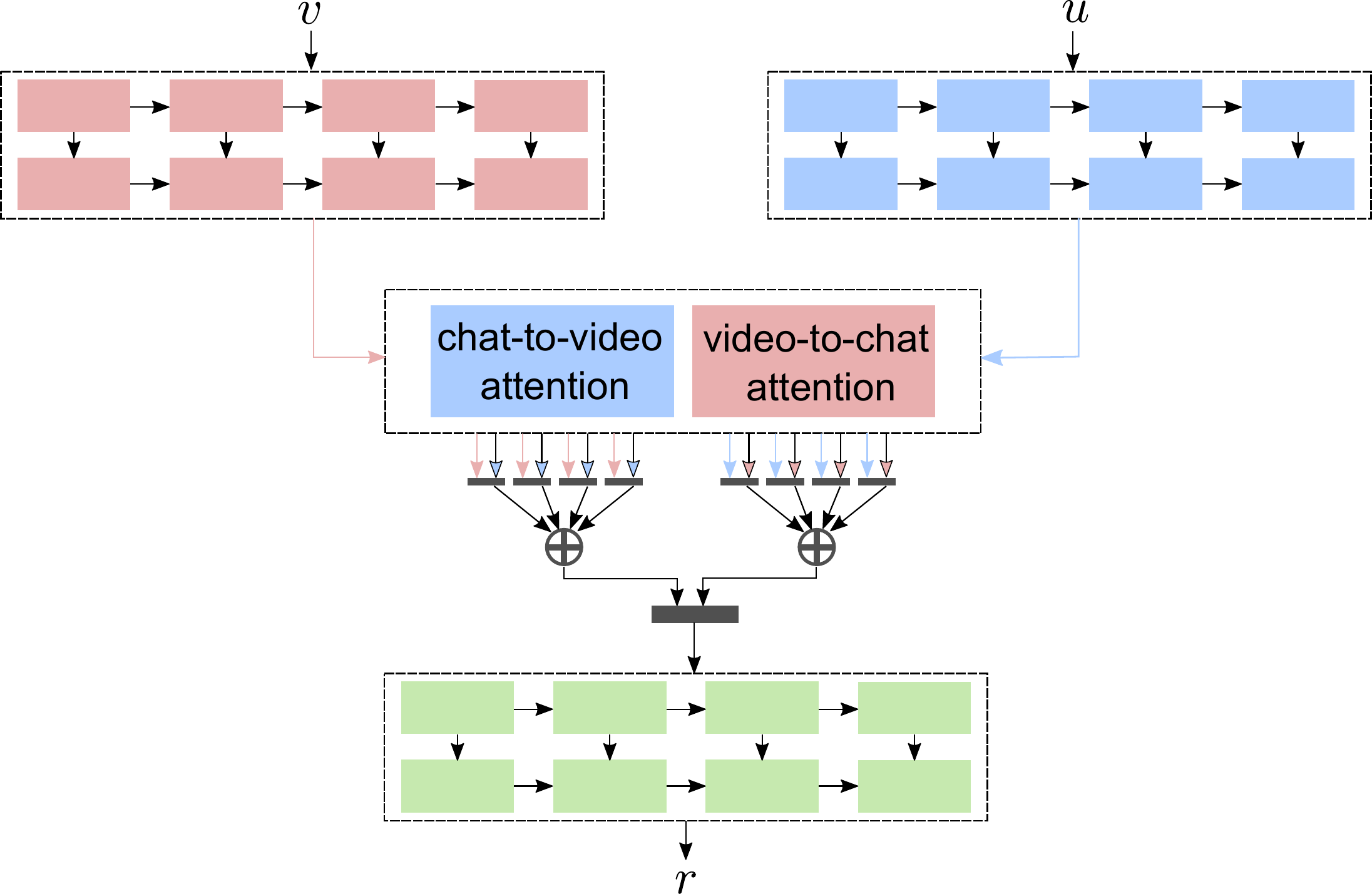}
\vspace{-9pt}
\caption{Overview of our generative model with bidirectional attention flow between video context and chat context during response generation. \label{fig:generative_model} \vspace{-7pt}}
\end{figure}

Our simpler generative model is a sequence-to-sequence model with bilinear attention mechanism (similar to~\newcite{luong2015effective}). 
We have two encoders, one for encoding the video context and another for encoding the chat context, as shown in Fig.~\ref{fig:generative_model}. We combine the final state information from both encoders and give it as initial state to the response generation decoder. The two encoders and the decoder are all two-layer LSTM-RNNs.
Let $h^v_i$ and $h^u_j$ be the hidden states of video and chat encoders at time step $i$ and $j$ respectively. At each time step $t$ of the decoder with hidden state $h^r_t$, the decoder attends to parts of video and chat encoders and uses the combined information to generate the next token. Let $\alpha_t$ and $\beta_t$ be the attention weight distributions for video and chat encoders respectively with video context vector $c^v_t =\sum_i \alpha_{t,i} h^v_{i}$ and chat context vector $c^u_t = \sum_{j}\beta_{t,j}h^u_{j}$. The attention distribution for video encoder is defined as (and the same holds for chat encoder):
\begin{align}
e_{t,i} &= {h^r_t}^T W^v_a h^v_i; \;\;\;\; \alpha_{t} = \mathrm{softmax}(e_{t})
\end{align}
where $W^v_a$ is a trainable parameter. 
Next, we concatenate the attention-based context information ($c^v_t$ and $c^u_t$) and decoder hidden state ($h^r_t$), and do a non-linear transformation to get the final hidden state ${\hat{h}^r_t}$ as follows:
\begin{equation}
\hat{h}^r_t = \tanh(W_c[c^v_t;c^u_t;h^r_t])
\end{equation}
where $W_c$ is again a trainable parameter. Finally, we project the final hidden state information to vocabulary size and give it as input to a \emph{softmax} layer to get the vocabulary distribution $p(r_t|r_{1:t-1},v,u;\theta)$. During training, we minimize the cross-entropy loss defined as follows:
\begin{equation}
L_{\mathrm{XE}}(\theta) = - \sum \sum_t \log p(r_t|r_{1:t-1},v,u;\theta)
\end{equation}
where the final summation is over all the training triples in the dataset. 

Further, to train a stronger generative model with negative training examples (which teaches the model to give higher generative decoder probability to the positive response as compared to all the negative ones), we use a max-margin loss (similar to~\eqnref{eq:max-margin-loss} in~\secref{subsec:triple-encoder}):
\begin{equation}
\small
\begin{aligned}
L_{\mathrm{MM}}(\theta) &= \sum [\max(0,M+ \log p(r|v',u)- \log p(r|v,u)) \\
& + \max(0,M+ \log p(r|v,u')- \log p(r|v,u)) \\
& + \max(0,M+ \log p(r'|v,u)- \log p(r|v,u))]
\end{aligned}
\end{equation}
where the summation is over all the training triples in the dataset. Overall, the final joint loss function is a weighted combination of cross-entropy loss and max-margin loss: $
L(\theta) = L_{\mathrm{XE}}(\theta) + \lambda L_{\mathrm{MM}}(\theta)$, where $\lambda$ is a tunable hyperparameter.

\subsubsection{Bidirectional Attention Flow (BiDAF)}
The stronger version of our generative model extends the two-encoder-attention-decoder model above to add bidirectional attention flow (BiDAF) mechanism~\cite{seo2016bidirectional} between video and chat encoders, as shown in Fig.~\ref{fig:generative_model}. Given the hidden states $h^v_i$ and $h^u_j$ of video and chat encoders at time step $i$ and $j$, the final hidden states after the BiDAF are $\hat{h}^v_i=[h^v_i;c^{v \leftarrow u}_i]$ and $\hat{h}^u_j=[h^u_i;c^{u \leftarrow v}_j]$ (similar to as described in Sec.~\ref{subsubsec:tri-daf}), respectively. Now, the decoder attends over these final hidden states, and the rest of the decoder process is similar to Sec~\ref{subsubsec:seq2seq} above, including the weighted joint cross-entropy and max-margin loss.

\section{Experimental Setup}
\label{sec:setup}
\paragraph{Evaluation}
\label{subsec:evaluation-metrics}
We first evaluate both our discriminative and generative models using retrieval-based recall@k scores, which is a concrete metric for such dialogue generation tasks~\cite{lowe2015ubuntu}. For our discriminative models, we simply rerank the given responses (in a candidate list of size 10, based on 9 negative examples; more details below) in the order of the probability score each response gets from the model. If the positive response is within the top-k list, then the recall@k score is $1$, otherwise $0$, following previous Ubuntu-dialogue work~\cite{lowe2015ubuntu}. For the generative models, we follow a similar approach, but the reranking score for a candidate response is based on the log probability score given by the generative models' decoder for that response, following the setup of previous visual-dialog work~\cite{das2016visual}. In our experiments, we use recall@1, recall@2, and recall@5 scores. 
For completeness, we also report the phrase-matching metric scores:
METEOR~\cite{denkowski2014meteor} and ROUGE~\cite{lin2004rouge}
for our generative models. We also present human evaluation.

\paragraph{Training Details} 
For negative samples, during training, for every positive triple (video, chat, response) in the training set, we sample $3$ random negative triples. For validation/test, we sample $9$ random negative responses elsewhere from the validation/test set. Also, the negative samples don't come from the video corresponding to the positive response. More details of negative samples and other training details (e.g., dimension/vocab sizes, visual feature details, validation-based hyperparamater tuning and model selection), are discussed in the supplementary.


\begin{table}
\begin{small}
\centering
\begin{tabular}{|l|c|c|c|}
\hline
Models & r@1 & r@2 & r@5\\
\hline
\multicolumn{4}{|c|}{\textsc{Baselines}}\\
\hline
Most-Frequent-Response & 10.0 & 16.0 & 20.9 \\
Naive Bayes & 9.6 & 20.9 & 51.5 \\
Logistic Regression & 10.8 & 21.8 & 52.5 \\
Nearest Neighbor  & 11.4 & 22.6 & 53.2 \\
Chat-Response-Cosine  & 11.4 & 22.0 & 53.2 \\
\hline
\multicolumn{4}{|c|}{\textsc{Discriminative Model}}\\
\hline
Dual Encoder (C)& 17.1 & 30.3 & 61.9 \\
Dual Encoder (V) & 16.3 & 30.5 & 61.1 \\
Triple Encoder (C+V) & 18.1 & 33.6 & 68.5 \\
TriDAF+Self Attn (C+V) & 20.7 & 35.3 & 69.4 \\
\hline
\multicolumn{4}{|c|}{\textsc{Generative Model}}\\
\hline
Seq2seq +Attn (C) & 14.8 & 27.3 & 56.6 \\
Seq2seq +Attn (V) & 14.8 & 27.2 & 56.7 \\
Seq2seq + Attn (C+V) & 15.7 & 28.0 & 57.0 \\
Seq2seq + Attn + BiDAF (C+V) & 16.5 & 28.5 & 57.7 \\
\hline
\end{tabular}
\vspace{-4pt}
\caption{Performance of our baselines, discriminative models, and generative models for recall@k metrics on our Twitch-FIFA test set. C and V represent chat and video context, respectively.
\label{table:retrieval_results}
 \vspace{-10pt}
}
\end{small}
\end{table}

\section{Results and Analysis}

\subsection{Human Evaluation of Dataset}

First, the overall human quality evaluation of our dataset (shown in Table~\ref{table:dataset-human-eval}) demonstrates that it contains 90\% responses relevant to video and/or chat context.
Next, we also do a blind human study on the recall-based setup (on a set of $100$ samples from the validation set), where we anonymize the positive response by randomly mixing it with $9$ tricky negative responses in the retrieval list, and ask the user to select the most relevant response for the given video and/or chat context. We found that human performance on this task is around $55\%$ recall@1, demonstrating that this 10-way-discriminative recall-based task setup is reasonably challenging for humans,\footnote{This relatively low human recall@1 performance is because this is a challenging, 10-way-discriminative evaluation, i.e., the choice comes w.r.t. 9 tricky negative examples along with just 1 positive example (hence chance-baseline is only 10\%). Note that these negative examples are an artifact of specifically recall-based evaluation only, and will not affect the more important real-world task of response generation (for which our dataset's response quality is 90\%, as shown in Table~\ref{table:dataset-human-eval}). Moreover, our dataset filtering (see Sec.~\ref{sec:datafilter}) also `suppresses' simple baselines and makes the task even harder.} but also that there is a lot of scope for future model improvements because the chance baseline is only 10\% and the best-performing model so far (see Sec.~\ref{sec:discrim-results}) achieves only 22\% recall@1 (on dev set), and hence there is a large $33\%$ gap. 

\subsection{Baseline Results}
Table~\ref{table:retrieval_results} displays all our primary results. We first discuss results of our simple non-trained and trained baselines (see Sec.~\ref{subsec:very-simple-baselines}). The `Most-Frequent-Response' baseline, which just ranks the 10-sized response retrieval list based on their frequency in the training data, gets only around 10\% recall@1.\footnote{Note that the performance of this baseline is worse than the random choice baseline (recall@1:10\%, recall@2:20\%, recall@5:50\%) because our dataset filtering process already suppresses frequent responses (see~\secref{sec:datafilter}), in order to provide a challenging dataset for the community.} Our other non-trained baselines: `Chat-Response-Cosine' and `Nearest Neighbor', which ranks the candidate responses based on (Twitch-trained RNN encoder's vector) cosine similarity with chat-context and $K$-best training contexts' response vectors, respectively, achieves slightly better scores. We also show that our simple trained baselines (logistic regression and nearest neighbor) also achieve relatively low scores, indicating that a simple, shallow model will not work on this challenging dataset.

\subsection{Discriminative Model Results}
\label{sec:discrim-results}
Next, we present the recall@k retrieval performance of our various discriminative models in Table~\ref{table:retrieval_results}: dual encoder (chat context only), dual encoder (video context only), triple encoder, and TriDAF model with self-attention. Our dual encoder models are significantly better than random choice and all our simple baselines above, and further show that they have complementary information because using both of them together (in `Triple Encoder') improves the overall performance of the model. Finally, we show that our novel TriDAF model with self-attention performs significantly better than the triple encoder model.\footnote{Statistical significance of $p < 0.01$ for recall@1, based on the bootstrap test~\cite{noreen1989computer,efron1994introduction} with 100K samples.}

\begin{table}
\centering
\small
\begin{tabular}{|l|c|c|}
\hline
Models & METEOR & ROUGE-L \\
\hline
\multicolumn{3}{|c|}{\textsc{Multiple References}}\\
\hline
Seq2seq + Atten. (C) &  2.59 & 8.44 \\
Seq2seq + Atten. (V) & 2.66 & 8.34 \\
Seq2seq + Atten. (C+V) $\otimes$ &  3.03 & 8.84\\
$\otimes$ + BiDAF (C+V) & 3.70 & 9.82 \\
\hline
\end{tabular}
\vspace{-3pt}
\caption{Performance of our generative models on phrase matching metrics.
\vspace{-1pt}}
\label{table:generative-mscoco-results}
\end{table}

\begin{table}[t]
\small
\centering
\begin{tabular}{l c c}
\hline
Models & Relevance \\
\hline
BiDAF wins & 41.0 \%  \\
Seq2seq + Atten. (C+V) wins & 34.0 \%  \\
Non-distinguishable & 25.0 \% \\
\hline
\end{tabular}
\vspace{-3pt}
\caption{Human evaluation comparing the baseline and BiDAF generative models. \label{table:generative-human-eval} \vspace{-6pt}}
\end{table}

\subsection{Generative Model Results}
\label{subsec:results-generative}
Next, we evaluate the performance of our generative models with both retrieval-based recall@k scores and phrase matching-based metrics as discussed in Sec.~\ref{subsec:evaluation-metrics} (as well as human evaluation). We first discuss the retrieval-based recall@k results in Table~\ref{table:retrieval_results}. Starting with a simple sequence-to-sequence attention model with video only, chat only, and both video and chat encoders, the recall@k scores are better than all the simple baselines. Moreover, using both video+chat context is again better than using only one context modality. Finally, we show that the addition of the bidirectional attention flow mechanism improves the performance in all recall@k scores.\footnote{Stat. signif. $p<0.05$ for recall@1 w.r.t. Seq2seq+Atten (video+chat); $p<0.01$ w.r.t. chat- and video-only models.} Note that generative model scores are lower than the discriminative models on retrieval recall@k metric, which is expected (see discussion in previous visual dialogue work~\cite{das2016visual}), because discriminative models can tune to the biases in the response candidate options, but generative models are more useful for real-world tasks such as generation of novel responses word-by-word from scratch in Siri/Alexa/Cortana style applications (whereas discriminative models can only rank the pre-given list of responses).

We also evaluate our generative models with phrase-level matching metrics: METEOR and ROUGE-L, as shown in Table~\ref{table:generative-mscoco-results}. Again, our BiDAF model is stat. significantly better than non-BiDAF model on both METEOR ($p<0.01$) and ROUGE-L ($p<0.02$) metrics. Since dialogue systems can have several diverse, non-overlapping valid responses, we consider a multi-reference setup where all the utterances in the $10$-sec response window are treated as valid responses.\footnote{\newcite{liu2016not} discussed that BLEU and most phrase matching metrics are not good for evaluating dialogue systems. Also, generative models have very low phrase-matching metric scores because the generated response can be valid but still very different from the ground truth reference~\cite{lowe2015ubuntu,liu2016not,li2016persona}. We present results for the relatively better metrics like paraphrase-enabled METEOR for completeness, but still focus on retrieval recall@k and human evaluation.}

\vspace{0pt}
\subsection{Human Evaluation of Models}
\vspace{0pt}
Finally, we also perform human evaluation to compare  our top two generative models, i.e., the video+chat seq2seq with attention and its extension with BiDAF (Sec.~\ref{subsec:generative-model}), based on a 100-sized sample.
We take the generated response from both these models, and randomly shuffle these pairs to anonymize model identity. We then ask two annotators (for $50$ task instances each) to score the responses of these two models based on relevance. Note that the human evaluators were familiar with Twitch FIFA-18 video games and also the Twitch's unique set of chat mannerisms and emotes.
As shown in Table~\ref{table:generative-human-eval}, our BiDAF based generative model performs better than the non-BiDAF one, which is already quite a strong video+chat encoder model with attention.

\begin{figure*}
\centering
\includegraphics[width=0.98\linewidth]{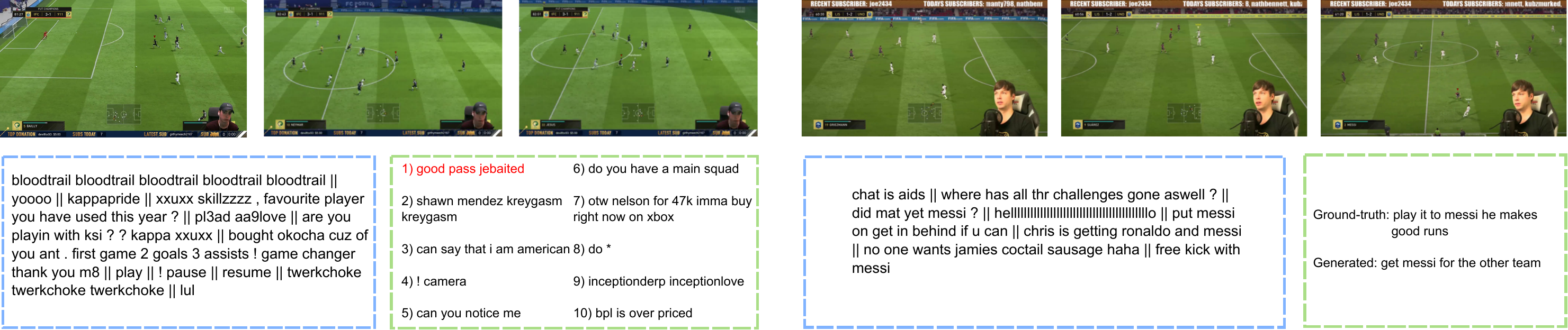}
\vspace{-9pt}
\caption{Output retrieval (left) and generative (right) examples from TriDAF and BiDAF models, resp. \label{fig:output-retrieval-examples} \vspace{7pt}}
\end{figure*}

\begin{figure*}
\centering
\includegraphics[width=0.98\linewidth]{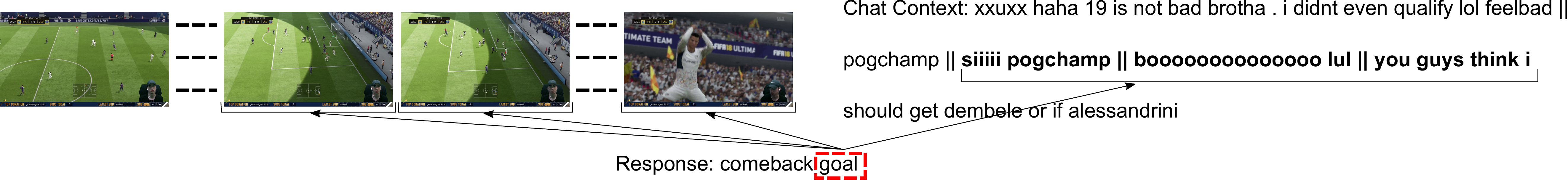}
\vspace{-8pt}
\caption{Attention visualization: generated word `goal' in response is intuitively aligning to goal-related video frames (top-3-weight frames highlighted) and context words (top-10-weight words highlighted).\vspace{-2pt}}
\label{fig:visualize-attention}
\end{figure*}

\begin{table}[t]
\small
\centering
\begin{tabular}{|l|c|c|c|}
\hline
Models & recall@1 & recall@2 & recall@5\\
\hline
1 neg.  & 18.21 & 32.19 & 64.05 \\
3 neg. &  22.20 & 35.90 & 68.09  \\
\hline
\end{tabular}
\vspace{-3pt}
\caption{Ablation (dev) of one vs. three negative examples for TriDAF self-attention discriminative model. \label{table:margin-loss-variants} \vspace{-7pt}}
\end{table}

\section{Ablations and Analysis}

\subsection{Negative Training Pairs}
We also compare the effect of different negative training triples that we discussed in Sec.~\ref{sec:setup}. Table~\ref{table:margin-loss-variants} shows the comparison between one negative training triple (with just a negative response) vs. three negative training triples (one with negative video context, one with negative chat context, and another with negative response), showing that using the 3-negative examples setup is substantially better.

\subsection{Discriminative Loss Functions}
Table~\ref{table:loss-function-comparison} shows the performance comparison between the classification loss and max-margin loss on our TriDAF with self-attention discriminative model (Sec.~\ref{subsubsec:tri-daf}). We observe that max-margin loss performs better than the classification loss, which is intuitive because max-margin loss tries to differentiate between positive and negative training example triples. 
\begin{table}[h]
\small
\centering
\begin{tabular}{|l|c|c|c|}
\hline
Models & recall@1 & recall@2 & recall@5\\
\hline
Classification loss  & 19.32 & 33.72 & 66.60 \\
Max-margin loss & 22.20 & 35.90 & 68.09  \\
\hline
\end{tabular}
\vspace{-2pt}
\caption{Ablation of classification vs. max-margin loss on our TriDAF discriminative model (on dev set).\vspace{-2pt}}
\label{table:loss-function-comparison}
\end{table}

\subsection{Generative Loss Functions}
For our best generative model (BiDAF), Table~\ref{table:loss-function-comparison-generative} shows that using a joint loss of cross-entropy and max-margin is better than just using only cross-entropy loss optimization (Sec.~\ref{subsubsec:seq2seq}). Max-margin loss provides knowledge about the negative samples for the generative model, hence improves the retrieval-based recall@k scores.
\begin{table}[h]
\small
\centering
\begin{tabular}{|l|c|c|c|}
\hline
Models & recall@1 & recall@2 & recall@5\\
\hline
Cross-entropy (XE)  & 13.12 & 23.45 & 54.78 \\
XE+Max-margin & 15.61 & 27.39 & 57.02 \\
\hline
\end{tabular}
\caption{Ablation of cross-entropy loss vs. cross-entropy+maxmargin loss for our BiDAF-based generative model (on dev set).
\vspace{-10pt}}
\label{table:loss-function-comparison-generative}
\end{table}

\subsection{Attention Visualization and Examples}
Finally, we show some interesting output examples from both our discriminative and generative models as shown in Fig.~\ref{fig:output-retrieval-examples}. Additionally, Fig.~\ref{fig:visualize-attention} visualizes that our models can learn some correct attention alignments from the generated output response word to the appropriate (goal-related) video frames as well as chat context words.

\vspace{-4pt}
\section{Conclusion}
\vspace{-4pt}
We presented a new game-chat based video-context, many-speaker dialogue task and dataset. We also presented several baselines  and state-of-the-art discriminative and generative models on this task. We hope that this testbed will be a good starting point to encourage future work on the challenging video-context dialogue paradigm. In future work, we plan to investigate the effects of multiple users, i.e., the multi-party aspect of this dataset. We also plan to explore advanced video features such as activity recognition, person identification, etc.

\vspace{-3pt}
\section*{Acknowledgments}
\vspace{-3pt}
We thank the reviewers for their helpful comments. This work was supported by DARPA YFA17-D17AP00022, ARO-YIP Award W911NF-18-1-0336, Google Faculty Research Award, Bloomberg Data Science Research Grant, and NVidia GPU awards. The views, opinions, and/or findings contained in this article are those of the authors and should not be interpreted as representing the official views or policies, either expressed or implied, of the funding agency.

\appendix
\vspace{15pt}
\noindent
{\fontsize{12}{12}\selectfont \textbf{Supplementary Material}} \par

\section{Simple Baselines}
\label{subsec:app-very-simple-baselines}
\subsection{Most-Frequent-Response}
As a simple non-trained retrieval baseline, we just return (or rank) the responses (in the retrieval ranking list; see supplementary Sec.~\ref{sec:app-setup}) based on their frequency in the training set. 
\subsection{Chat-Response-Cosine}
We also use another simple non-trained baseline following previous work~\cite{lowe2015ubuntu}, where we choose/rank the candidate responses based on the cosine similarity between the vector representations of the given chat context and each candidate response (in the retrieval list). We train an LSTM-RNN language model on the Twitch training chat context data. We then use the final hidden state of this pretrained RNN to represent the given chat context/response.
\subsection{Nearest Neighbor}
We also use a nearest neighbor non-trained baseline similar to~\newcite{das2016visual}, where, given the chat context, we find the $K$-best similar chat contexts in the training set and take their corresponding responses. Next, we again rerank the candidate responses based on the mean similarity score between the candidate response and these $K$ nearest-neighbor responses. Here again, we use a pretrained Twitch-LM RNN for vector representation of chat contexts and responses.

\subsection{Logistic Regression and Naive Bayes}
Apart from the non-trained baselines, we also present simple trained baselines based on logistic regression and Naive Bayes. Here again, we use the pretrained RNN for chat context and response vector representations.

\section{Experimental Setup}
\label{sec:app-setup}

\subsection{Evaluation}
\label{subsec:app-evaluation-metrics}
We first evaluate both our discriminative and generative models using retrieval-based recall@k scores, which is a concrete metric for such dialogue generation tasks~\cite{lowe2015ubuntu}. For our discriminative models, we simply rerank the given responses (in a candidate list of size 10, based on 9 negative examples as described below) in the order of the probability score each response gets from the model. If the positive response is within the top-k list, then the recall@k score is $1$, otherwise $0$, following previous Ubuntu-dialogue work~\cite{lowe2015ubuntu}. For the generative models, we follow a similar approach, but the reranking score for a candidate response is based on the log probability score given by the generative models' decoder for that response, following the setup of previous visual-dialog work~\cite{das2016visual}. In our experiments, we use recall@1, recall@2, and recall@5 scores. 
For completeness, we also report the phrase-matching metric scores:
METEOR~\cite{denkowski2014meteor} and ROUGE~\cite{lin2004rouge}
for our generative models. Because dialogue models are hard to evaluate using such phrase-matching metrics~\cite{liu2016not}, we also perform human evaluation based comparison for our two strongest generative models (Seq2seq+attention and Seq2seq+attention+BiDAF).

\begin{figure*}[t]
\centering
\includegraphics[width=0.98\linewidth]{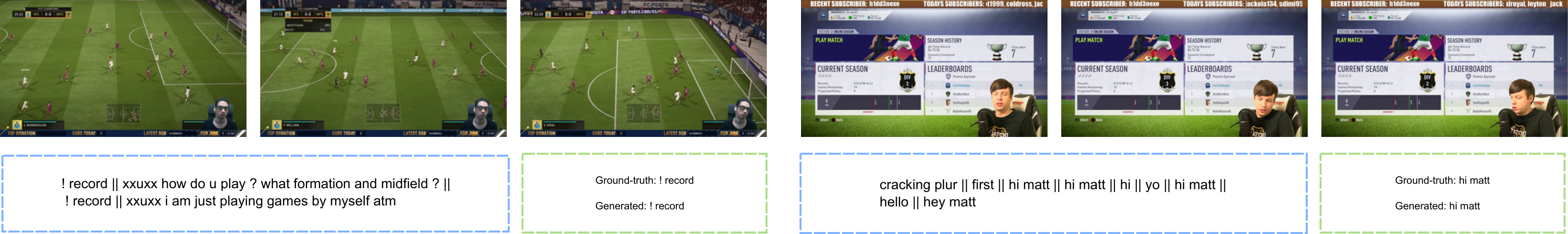}
\vspace{-3pt}
\caption{Output generative examples from our BiDAF model.\vspace{-1pt}}
\label{fig:output-extra-examples}
\end{figure*}

\subsection{Negative Samples (Training and Val/Test)}
\label{subsec:app-setup-negative-examples}
Both our discriminative and generative models need negative samples during training and as well as for the test time (dev/test) retrieval lists for recall@k scores. For training, for every positive triple (video, chat, response) in the training set, we sample $3$ random negative triples elsewhere from the training set such that the negative sample triples do not come from the video corresponding to the positive triple. We refer Sec. 4.2.1 (of main paper) for details about how we add these negative samples in the training. For validation/test, we sample $9$ random negative responses elsewhere from the validation/test set for the given video and chat context, such that they don't come from the video corresponding to the positive response, so as to create a 10-sized retrieval list.

\subsection{Training Details}
\label{app-subsec:training-details}
For tokenizing the word-level utterances in the chat context and response, we use the Twitter Tokenizer from NLTK library\footnote{\url{http://www.nltk.org}}. All the video clips are down-sampled to $3$ fps (frames per second) and also cropped to $244 \times 244$ size. Further, we extract the inception-v3~\cite{szegedy2016rethinking} frame-level features (standard penultimate layer) of $2048$ dimension from each video clip and use it as input to our video context encoder.
For all of our models, we tune the hyperparameters on the validation set. Our model selection criteria is based on recall@1 for discriminative model and METEOR for the generative model. We set the hidden unit size of $256$ dimension for LSTM-RNN. We down-project the inception-v3 frame level features of size $2048$ to $256$ dimension before feeding it as input to the video context encoder. We use word-level RNN model with $100$ dimension word embedding size and a vocabulary size of $27,000$. We initialize the word embeddings with Glove~\cite{pennington2014glove} vectors. We unroll the video context LSTM to a maximum of $60$ time steps and the chat context to a maximum of $70$ time steps. For the response, we unroll the RNN to a maximum of $10$ time steps. All of our models use Adam~\cite{kingma2014adam} optimizer with a learning rate of $0.0001$ (unless otherwise specified) and a batch size of $32$. Also, we use gradient clipping with maximum clip norm value of $2.0$. We set the margin $M$ in our max-margin loss to $0.1$ for all models. We use $\lambda=1.0$ for the weighted loss in the generative model.

\section{Output Examples}
Fig.~\ref{fig:output-extra-examples} presents additional examples for output responses generated by our BiDAF model.

\bibliography{citations}
\bibliographystyle{acl_natbib_nourl}

\end{document}